\documentclass[letterpaper, 10 pt, conference]{ieeeconf}  % Comment this line out
\IEEEoverridecommandlockouts                              % This command is only
\usepackage[utf8]{inputenc}
\usepackage[T1]{fontenc}
\usepackage{xcolor}
\usepackage{hyperref}

\usepackage{graphicx} % for pdf, bitmapped graphics files
\usepackage{amsmath} % assumes amsmath package installed

\title{\LARGE \bf
Biologically Inspired Model for Timed Motion in Robotic Systems
}

\author{Sebastian Doliwa$^{1}$ and Muhammad Ayaz Hussain$^{2}$ and Tim Sziburis$^{3}$ and Ioannis Iossifidis$^{4}$% <-this % stops a space
\thanks{This work is supported by the Ministry of Economics, Innovation, Digitization and Energy of the State of North Rhine-Westphalia and the European Union, grants GE-2-2-023A (REXO) and IT-2-2-023 (VAFES)}% <-this % stops a space
\thanks{$^{1}$ Sebastian Doliwa is member of the Iossifidis-Lab and  with the Department of Computer Science, Ruhr West University of Applied Sciences, 45407 Mülheim an der Ruhr, Germany {\tt\small sebastian.doliwa@hs-ruhrwest.de}}%
\thanks{$^{2}$ Muhammad Ayaz Hussain is member of the Iossifidis-Lab and with the Department of Computer Science, Ruhr West University of Applied Sciences, 45407 Mülheim an der Ruhr, Germany {\tt\small muhammad.hussain@hs-ruhrwest.de}}%
\thanks{$^{3}$ Tim Sziburis is member of the Iossifidis-Lab and  with the Department of Computer Science, Ruhr West University of Applied Sciences, 45407 Mülheim an der Ruhr, Germany {\tt\small tim.sziburis@hs-ruhrwest.de}}%
\thanks{$^{4}$ Ioannis Iossifidis is head of the Iossifidis-Lab and  with the Department of Computer Science, Ruhr West University of Applied Sciences, 45407 Mülheim an der Ruhr, Germany {\tt\small iossifidis@hs-ruhrwest.de}}%
}

\begin{document}

\maketitle
\thispagestyle{empty}
\pagestyle{empty}

%%%%%%%%%%%%%%%%%%%%%%%%%%%%%%%%%%%%%%%%%%%%%%%%%%%%%%%%%%%%%%%%%%%%%%%%%%%%%%%%
\begin{abstract}
The goal of this work is the development of a motion model for sequentially timed movement actions in robotic systems under specific consideration of temporal stabilization, that is maintaining an approximately constant overall movement time (isochronous behavior). This is demonstrated both in simulation and on a physical robotic system for the task of intercepting a moving target in three-dimensional space.

Motivated from humanoid motion, timing plays a vital role to generate a naturalistic behavior in interaction with the dynamic environment as well as adaptively planning and executing action sequences on-line.
%In humanoid movements, the timing of motion and action sequences plays a vital role in achieving desirable properties such as temporal stabilization.
In biological systems, many of the physiological and anatomical functions follow a particular level of periodicity and stabilization, which exhibit a certain extent of resilience against external disturbances. A main aspect thereof is stabilizing movement timing against limited perturbations. Especially human arm movement, namely when it is tasked to reach a certain goal point, pose or configuration, shows a stabilizing behavior.

%The goal of this work is to demonstrate the characteristics of temporal stabilization for sequentially timed movement actions on a machine.
%In this work, an anthropomorphic robot arm is used to intercept a moving target in 3D space. 
This work incorporates the utilization of an extended Kalman filter (EKF) which was implemented to predict the target position while coping with non-linear system dynamics.
The periodicity and temporal stabilization in biological systems was artificially generated by a Hopf oscillator, yielding a sinusoidal velocity profile for smooth and repeatable motion.
\end{abstract}
\begin{keywords}
test
\end{keywords}

%%%%%%%%%%%%%%%%%%%%%%%%%%%%%%%%%%%%%%%%%%%%%%%%%%%%%%%%%%%%%%%%%%%%%%%%%%%%%%%%
\section{Introduction}
Modeling movement and trajectory generation of biological systems can be addressed on different abstraction scales. These conceptional levels comprise, i.~a., nervous and muscular systems.
In this work we focused on mimicking aspects of these systems on a very basic level and transferring human behaviour to a robotic machine, in particular the reproducible timing of sequential motion actions under consideration of isochronous temporal stabilization.

Unlike traditional approaches which are often predetermined, the followed concept was designed to behave adaptively. This means, that -- solely based on visual perception -- the robot is capable to fulfill its task of intercepting a moving pendulum. Furthermore, the total movement time is independent of the desired destination, which is not the case in classical approaches.

Dynamical systems in general can be used to produce oscillations in phase space, where parameters are adjustable in real-time in order to achieve a specific goal. For instance, the oscillator radius can be adjusted to change the resulting velocity for matching a set temporal criterion. Additionally changing the parameters can make a limit cycle emerge or disappear. This feature can be used to let the system stay at a specific state.

For enabling a robot to perceive objects, a visual object detection algorithm is used in this work to keep track of a moving target. Once recognized, the object's current position is sent to an EKF.

The EKF stems from the field of statistics and control theory. It uses system and noise models for reducing uncertainty and predicting system states. Inside the EKF, the system variables, such as the system states and the corresponding noises, are constantly updated. Essential requirement for the utilization of this filter is that the behavior of the considered system can be mathematically described. By integrating the EKF into a processing chain, the exact current and future system states can be derived.

Following these calculations, an interception condition is constantly checked, and once fulfilled, a Hopf oscillator -- based on dynamical system theory -- is started. This isochronous limit cycle is used to create a sinusoidal velocity profile and to act as a temporal stabilizer of the whole movement generation. The robot arm follows a trajectory generated by a dynamical system and for each position, the inverse kinematics is calculated which yields the particular angle for every joint.

Afterwards, using the robot's network connection, the calculated joint angles are transmitted to the robotic system. 
To verify the trajectory generation, the dynamical system implementation as well as the inverse kinematics, a simulation environment was created in Unity.

As simulations of the whole system could promisingly confirm the proposed behavior, the algorithm was subsequently implemented on a Kuka LBR iiwa 7 R800 anthropomorphic robot arm where it also proved successful in providing temporally stabilized timed motion sequences.

With that, the goal to mimic smooth movement and trajectory generation of biological systems on a machine could be achieved by making use of object detection, an EKF for predicting future positions, and incorporating dynamical system theory to model the isochronous behaviour for timed motion in robotic systems. This combination of mathematical tools enabled us to transfer actions which usually can be exerted intuitively by human-beings to a machine.

\section{Related Work}
%The generation of trajectories inspired from biology is coped with in \cite{Flash}.

There is sizeable literature on the planning and control of human arm movements which is summarized by Flash et al. \cite{Flash}, reflecting sophisticated exact approaches. They specifically mention the empirical two thirds power law which describes the relation between trajectory curvature and angular velocity. Supposedly stemming from the central nervous system, it yields the velocity as piecewise proportional from the curvature to the power of two thirds \cite{Lacquaniti1983Oct}. As an additional principle they highlight concepts of optimization, such as minimum jerk, minimum torque change, minimum acceleration or minimum variance to account for smoothness of generated point-to-point and obstacle avoidance movements. They point out the relation between two-thirds power law and optimization as well as the supposed coupling between geometry and temporal aspects of movement. Geometrical transformation were also used to describe movement generation from sets of invariant primitives. They state, while ``equi-affine geometry successfully accounts for the two-thirds power law it cannot account for isochrony'' \cite{Flash} (affine transformations may cope with that).
Furthermore, these approaches do not provide an explicit stabilization of movements.

Apart from that, in a specifically temporal perspective with regard to isochrony and bell-shaped velocity profiles as observed in natural movements \cite{Grimme2012Oct}, neurobiologically motivated notions based on central pattern generators (CPGs), inspired from the nervous system and spinal cord \cite{Matsuoka1987}, have proven successful \cite{Ijspeert2008}.
Non-linear oscillators can be interpreted in this regard for timed movement sequence generation and establishing isochronous, stabilized motion. In the simulation studies of \cite{Schoner} this has been shown for the interception of moving objects by a two-degrees-of-freedom (DoF) robotic arm, and the temporal coordination of two 6-DoF arms. Moreover, this concept has for example been applied to continuously hitting a ball on an inclined plane while utilizing linear Kalman filtering \cite{Oubbati2013}, and catching a ball moving on a table \cite{Santos2009}. As these tasks are the ones which are primarily related to the use-case examined in this work, they will be explained in the following.

Oubatti et al. \cite{Oubbati2013} accomplish the coordination of a number of differently timed autonomous actions with a ``behavioral organization architecture that is sensitive to timing''. They incorporate two switchable movement regimes which establish the end-effector motion from a current pose state to a target state within a defined duration (isochrony), when combining them. The pose state implements a regime with a fixed point attractor, while a Hopf oscillator is applied for an oscillatory regime as stabilization for ``a periodic solution along a limit cycle attractor''. Besides these regimes, a robust motion is generated by considering Gaussian white noise in the movement dynamics to ``escape from unstable states''. Since the target state prediction can vary over time, the movement parameterization is continuously adapted, also comprising the cycle time for de- and acceleration respectively.

A similar approach was followed by Santos and Ferreira \cite{Santos2009}, who differentiated between and combined three movement regimes, characterizing the initial state and the final state as discrete components, together with a harmonic limit cycle as stable oscillatory intermediate component. While the discrete components are represented by globally attractive fixed points, the limit cycle solution is generated by a Hopf oscillator as a bifurcation from a fixed point. Again, the movement dynamics are extended by a Gaussian white noise term. In turn, a continuous parameter adaption can be made use of for modulating the trajectory to be generated on-line.

Apart from that, the usage of dynamical systems for creating timely coordinated movements proved successful in a robotic context, i.~a., for performing drumming tasks \cite{Degallier2006}, discrete movement of a mobile robot under variable perturbations \cite{Tuma2009}, and controlling specific variants of gait in quadruped \cite{Santos2011} and snake-like robots \cite{Wang2016} or biologically inspired flight \cite{Chung2009}. With respect to biological inspiration, the trajectories generated with this approach have demonstrated to reflect characteristic properties of human motion \cite{Rano}. While being based on these concepts, our approach extends them by using an extended Kalman filter for prediction and correction to cope with non-linear system dynamics.

When looking into the concrete problem to enable anthropomorphic robot arms to exert movements for striking or catching ball-like objects, it can be seen this task has been addressed in a variety of work, whereby biologically inspired approaches comprise the generation of movement sequences under specific consideration of their timing, although the previously introduced approach of dynamic system limit cycles has not been applied yet, to the authors' knowledge.

The state-of-the-art workflow for the exemplarily introduced case consists of visual object tracking, prediction of its trajectory, followed by inverse kinematics calculations for the robot in order to plan and establish an appropriate trajectory, cf. \cite{Huang2015}.

Biomimetic approaches have been followed with the purpose of learning striking movements. Using stereo camera vision with an extended Kalman filter and a 7-DOF arm,
%With making use of an extended Kalman filter,
\cite{Mulling2010} splits such movements into four stages and introduces virtual hitting points. They propose to prospectively replace their approach of spline-based trajectories by motor primitives in a dynamical systems context -- which was eventually realized in \cite{Mulling2013} with kinesthetic teach-in for imitation learning. The imitation learning concept was followed in \cite{Park2009}, too, in the form of a genetic/evolutionary algorithm stemming from biological principles, exploring movement primitives reduced by principal component analysis, applied to a 6-DOF arm and using a depth camera with Kalman filtering.
Furthermore benefiting from optimization on the basis of natural human behavior, another approach distinguished between two strategies of playing (defensive and focused respectively) under consideration of joint limits and applying third order polynomial trajectories 
\cite{Koc2018}.
As a further approach towards timing, \cite{Hsiao2014} introduced a robotic system which is able to bat a ball to a desired position by also optimizing temporal aspects of the rebounding ball's arrival.

Although these approaches consider bio-imitation in the present use-case, they neither cope with an explicit stabilization of movements nor with the characteristic of isochrony.

%The approaches of using an extended Kalman filter and curve fitting for trajectory prediction were compared in \cite{Tebbe2019}, where a table tennis experiment was implemented on a 6-DOF robot arm. %, utilizing four cameras.
A comparison between applying a linear and a non-linear model with the respective Kalman filter implementations in a ball catching and juggling task was made in \cite{Rapp2011}, showing that ``shortcomings of the robots must sometimes be equalized by improving models to capture reality more closely'' -- which speaks in favor of using an extended Kalman filter and more general a biologically motivated model, inspired from naturally occurring systems.% (incorporating two cameras).

%Visual tracking by means of difference images with data coming from two cameras was used in \cite{Frese2001} for ball catching, integrating an extended Kalman filter for target tracking and prediction. \cite{Modi2005} utilizes a comparable visual technique with only one camera in the context of a robotic arm playing table tennis. For the same purpose, \cite{Yu2013} with simple curve fitting, as well as \cite{Li2012} with a two-phase curve fitting approach (rough prediction before bouncing; precise prediction after bouncing) for the ball, make use of two single cameras to be synchronized for contour shape feature detection, finally applied on a 7-DOF arm.

%When extending the scope from robot arms to general robotic systems, works like \cite{Angel2005}, \cite{Silva2005} and \cite{Trasloheros2014} present the application of a parallel robot for playing table tennis on a real-time system, making use of a camera on the end-effector and Kalman filtering, too. 

\section{Theoretical Considerations}

\begin{figure*}[!t]
    \centering
    \includegraphics[width=0.7\textwidth]{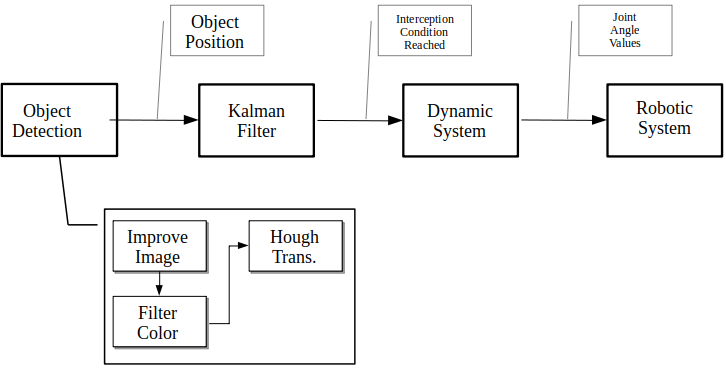}
    \caption{Structure of the overall project. At first, there is a simple combination of image processing algorithms giving the objects position, followed by a Kalman filter to correct and predict the object's position. When the interception condition is fulfilled, the Hopf oscillator is started to generate a sinusoidal velocity profile and to temporally stabilize the arms movement.}
    \label{fig:1}
\end{figure*}

%In the 1960s, NASA was looking for an algorithm for the correction of redundant data and prediction of system states. The result was the Kalman filter invented by Rudolf E. Kálmán, Richard S. Bucy und Ruslan L. Stratonovich, which was first used for the moon landing.
The Kalman filter is an algorithm which can be used to rectify redundant measurements, estimate and predict unknown system states.
The core of this filter is the state-space representation of a system. Estimates are based on this model and the known input variables. In order to minimize the estimation error, the optimal control method is used in the design of the Kalman filter instead of the pole specification method.
The original specification works solely for linear systems because the estimation is based on a Gauss contribution which is only applicable to linear systems. Since our work is based on a non-linear system, an extension -- the EKF -- is needed. In order to linearize the system, finite differences are applied to the system matrices.

Henri Poincaré developed the theory that after some time the system states are very close to their initial conditions. The mathematical definition of dynamical systems is as follows: A dynamical system is described by a triple $(T,X,\varphi^t)$. The parameter $T$ stands for the time, $X$ for the system states and $\varphi^t$ is the systems evolution vector $t \in T$, which is the main part. It describes the system evolution based on its initial conditions and can be used for the system's future approximation.
This means, the system's evolution only depends on its initial conditions and not only initial time.

\subsection{Structure and equations of the Kalman filter} The Kalman filter equations \eqref{eq:one}, \eqref{eq:two}, \eqref{eq:three}, \eqref{eq:four}, \eqref{eq:five}, \eqref{eq:six} can be derived from the structure (\autoref{fig:2}). The whole filter is based on the following formulation:

\begin{figure}[htp]
    \centering
    \includegraphics[width=8cm]{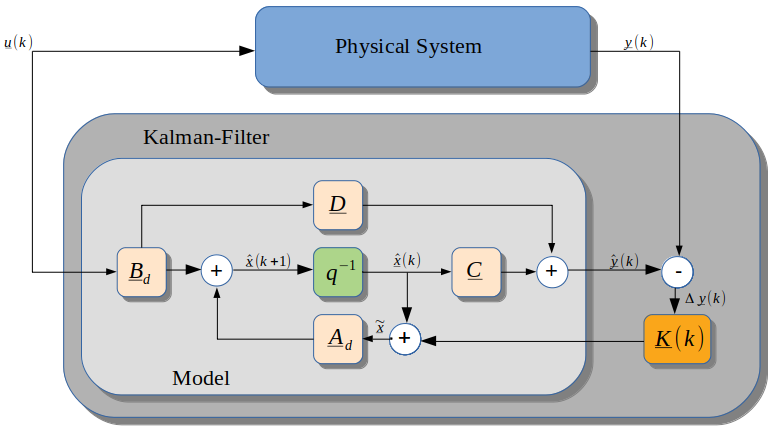}
    \caption{Structure of Kalman filter}
    \label{fig:2}
\end{figure}

Equations for prediction:
\begin{equation}
\label{eq:one}
\underline{\hat{x}}(k+1)=\underline{A}_d \cdot \underline{\tilde{x}}(k) + \underline{B}_d \cdot \underline{u}(k)
\end{equation}

\begin{align}
\label{eq:two}
\underline{\hat{P}}(k+1)&= \underline{A}_d \cdot \underline{\tilde{P}}(k) \cdot \underline{A}_d^T  + \underline{G}_d \cdot \underline{Q}(k) \cdot \underline{G}_d^T 
\end{align}

Equations for correction:

\begin{equation}
\label{eq:three}
\underline{K}(k)=\underline{\hat{P}}(k) \cdot \underline{C}^T \cdot (\underline{C} \cdot \underline{\hat{P}}(k) \cdot \underline{C}^T + \underline{R}(k) )^{-1}
\end{equation}

\begin{equation}
\label{eq:four}
\underline{\tilde{x}}(k+1)=\underline{\hat{x}}(k)+\underline{K}(k) \cdot ( \underline{y}(k) - \underline{C} \cdot \underline{\hat{x}}(k)-\underline{D} \cdot \underline{u}(k) )
\end{equation}

\begin{equation}
\label{eq:five}
\underline{\tilde{P}}(k+1)=( \underline{I}-\underline{K}(k) \cdot \underline{C}) \cdot \underline{\hat{P}}(k)
\end{equation}

\begin{equation}
\label{eq:six}
\Delta \underline{y} (k)= \underline{y}(k) - \underline{\hat{y}}(k)
\end{equation}

Equations \ref{eq:one} and \ref{eq:two} predict system states and their covariances. During the second step, the output value $\Delta y(k)$ (\ref{eq:six}) is calculated, which is the difference between the actual output value $y(k)$ and the predicted one $\hat{y}(k)$. This difference constitutes the Kalman gain $K(k)$ (\ref{eq:three}). For a high Kalman gain, the filter algorithm relies more on the actual measurements $y(k)$ than on its own system model. It follows that the system states $x(k)$ and their covariances $P(k)$ (\ref{eq:five}) have to be corrected. If the Kalman gain is low, the filter uses its system model and measurements have less impact on the system's state estimation. This is conducted periodically and in real-time.

\subsection{Dynamical Systems} In this work a limit cycle is used to maintain a constant movement time. A limit cycle is a periodic and isolated solution of a differential equation system in an equilibrium. These limit cycles are closed curves in phase space, which can attract trajectories (called attractor), or repel them (called repellor). It is also possible that trajectories are semi-stable, i.~e. running next to each other.
To keep control of this limit cycle, a Hopf bifurcation \eqref{eq:seven} is used. A bifurcation in general is a branch of systems which depend on parameters. 

\begin{equation}
\label{eq:seven}
    \dot{x} = f(x,\alpha)
\end{equation}

where $x$ is the systems state and $\alpha$ the parameter which controls the topology.

The special aspect about a Hopf bifurcation is that, depending on $\alpha$, a limit cycle arises or disappears.

\section{Methodology}
The whole system's structure can be seen in \autoref{fig:1}. First, there is the object recognition. Second, the Kalman filter which starts the dynamical system when the interception condition is fulfilled. Finally, the velocity profile generated by the dynamical system is executed on the robotic system.

\subsection{Kalman filter calculation} 
\begin{figure}[htp]
    \centering
    \includegraphics[width=4cm]{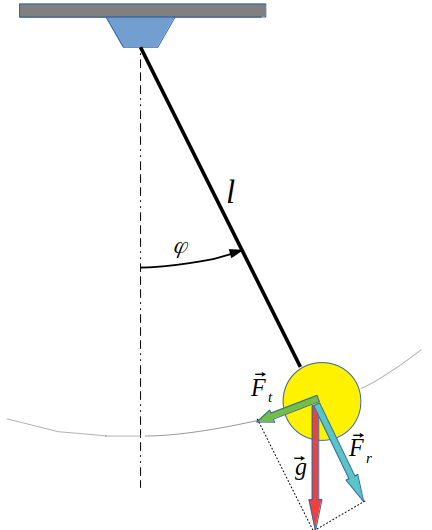}
    \caption{Pendulum forces}
    \label{fig:3}
\end{figure}

For the implementation of an EKF the system's motion model, the measurement noise, the system noise and an initial condition are needed. If the initial condition is very close the real one, the filter's prediction and correction of system values are more accurate from the beginning. The system in consideration is a damped pendulum. Based on \autoref{fig:3}, the forces and the resulting state space model (\autoref{eq:8}) can be derived.

\begin{equation}
\label{eq:8}
\begin{pmatrix}
\dot{\varphi}\\ 
\ddot{\varphi} 
\end{pmatrix}
=
\begin{pmatrix}
\dot{\varphi} \\
-\frac{g}{l} \sin(\varphi)-\alpha \dot{\varphi}
\end{pmatrix}
\end{equation}

However, this system model has to be discrete. \autoref{eq:9} represents the resulting system matrix.

\begin{align}
\label{eq:9}
f&=\vec{x}_k=
    \left[ \begin{matrix}
\varphi_k \\ 
\dot{\varphi_k}
\end{matrix} \right]=\\ \nonumber
&=\left[ \begin{matrix}
\varphi_{k-1} + T \dot{\varphi}_{k-1} + \frac{T^2}{2} \left(-\alpha \dot{\varphi}_{k-1}-\frac{g}{l} \cdot sin(\varphi_{k-1}) \right) \\
\dot{\varphi}_{k-1} + T \left(-\alpha \dot{\varphi}_{k-1}-\frac{g}{l} \cdot sin(\varphi_{k-1})\right) + n_1
\end{matrix} \right] 
\end{align}

Besides the discrete system model, the measurement matrix (\ref{eq:10}) is required. For the lateral measurement of the pendulum the following measurement matrix is derived:

\begin{equation}
\label{eq:10}
\vec{z}_k =
\left[ \begin{matrix}
x
\end{matrix} \right]
=
\left[ \begin{matrix}
l \cdot sin(\varphi_k) + v_1
\end{matrix} \right]
\end{equation}

Applying linearization to the system matrix $A$, input matrix $B$, measurement matrix $C$ and the state transition matrix $D$, the following results are obtained:

\begin{equation}
A_{[i,j]} = \frac{\vartheta f_{[i]}}{\vartheta x_{[j]} } =
\left[ \begin{matrix}
1- \frac{T^2g}{2l} \cdot cos(\varphi_{k-1}) & T-\frac{T^2}{2}\cdot \alpha \\
-\frac{Tg}{l} \cdot cos(\varphi_{k-1}) & 1-\alpha \cdot T
\end{matrix} \right]
\end{equation}

\begin{equation}
D_{[i,j]} = \frac{\vartheta f_{[i]}}{\vartheta n_{[j]} } =
\left[ \begin{matrix}
0 \\
1
\end{matrix} \right]
\end{equation}

\begin{equation}
C_{[i,j]} = \frac{\vartheta z_{[i]}}{\vartheta x_{[j]} } =
\left[ \begin{matrix}
cos(\varphi_k) & 0
\end{matrix} \right]
\end{equation}

\begin{equation}
V_{[i,j]} = \frac{\vartheta z_{[i]}}{\vartheta v_{[j]} } =
\left[ \begin{matrix}
1
\end{matrix} \right]
\end{equation}

Essential aspect of developing a Kalman filter is the correct estimation of the measurement noise matrix $R$ \eqref{eq:fifteen} and the system noise matrix $Q$ \eqref{eq:sixteen}. 

\begin{equation}
\label{eq:fifteen}
    R = [1000^2]
\end{equation}

The matrices $Q$ and $R$ are decisive for the filter behaviour and its quality. However, the correct estimation of these matrices is complex. For example, if a huge amount of measurement noise is guessed, the filter responds very sluggish to disturbances.
The following formula is used for calculating the system noise matrix $Q$:

\begin{equation}
\label{eq:sixteen}
\underline{Q}=\underline{G}_d \cdot \underline{G}^T_d \cdot \sigma
\end{equation}

where $\sigma$ represents the angular standard derivation. $G_d$ is the vector containing the systems noise and is calculated by:

\begin{equation}
\underline{G}_d = \int_{0}^{dT} e^{\underline{A} \cdot v} \cdot \underline{G} \hspace{1mm} dv
\end{equation}

The vector $G$ gives an insight on how the disturbances are affecting each system state. In this project, disturbances can only affect the pendulum's velocity according to this equation:

\begin{equation}
s = \frac{1}{2} \cdot a \cdot t^2
\end{equation}

which results the following statement for $G_d$:

\begin{equation}
\underline{G}_d = \left(
\begin{matrix}
\frac{1}{2} \cdot T^2 \\
T
\end{matrix} \right)
\end{equation}

The measurement noise is input into the matrix $R$ \eqref{eq:fifteen}. Since only one system state is measured, only one value is entered here. This value is difficult to derive precisely and was therefore determined experimentally. The value must be selected sufficiently large so that the filter is only slightly affected by interference.

\subsection{Simulation in MATLAB} 
\begin{figure}[htp]
    \centering
    \includegraphics[width = 8 cm]{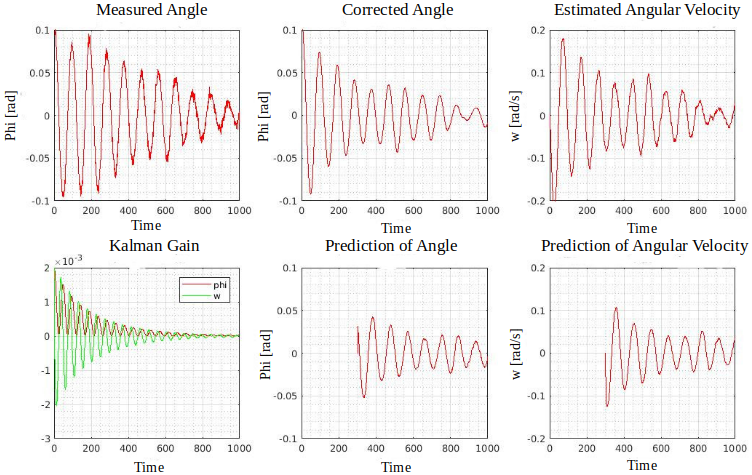}
    \caption{The plot ``Measured Angle'' contains the real measured angle. These values are input into the Kalman filter which results in the top two plots. Although only the pendulum's angle is measured, the Kalman filter is also giving an estimate for the angular velocity. The plots ``Prediction of Angle'' and ``Prediction of Angular Velocity'' contain the filter's prediction of 300 time steps (one time step is one iteration cycle within MATLAB, i.~e. the y axis represents the number of iteration cycles) for the pendulum's angle and angular velocity. The Kalman gain is shown in the bottom left plot. It is very low from the beginning and reduces to almost zero after 600 time steps.}
    \label{fig:four}
\end{figure}

In order to fit the EKF to the task, the system was simulated in MathWorks MATLAB\textsuperscript{\textregistered}. At first, the pendulum's position and the corresponding angles of ten complete oscillations were recorded. The simulation gave the opportunity to estimate the system and measurement noise. After tuning these parameters, \autoref{fig:four} shows the optimal working EKF.

\subsection{Dynamical Systems} The speed dynamics are based on the variables $\alpha$ (velocity) and an auxiliary variable $b$:

\begin{align}
\begin{split}
\tau \left(\begin{matrix}
\dot{a}\\
\dot{b}
\end{matrix}  \right) 
&= - c_1 \cdot u_{init}^2 
\left(\begin{matrix}
a\\
b
\end{matrix} \right) \\
&+ u_{Hopf}^2 \cdot f_{Hopf}(a-R_h, b) \\
&- c_2 \cdot u_{final}^2
\left(\begin{matrix}
a^2 -a \cdot \alpha_{tc}\\
b
\end{matrix} \right)
\end{split}
\end{align}

where $c_1,c_2$ are scaling parameters, $\alpha_{tc}$ the Hopf parameter and $f_{hopf}$ the Hopf oscillator \eqref{eq:21} with the radius $R_h$ \eqref{eq:23}.
The parameters $u_{init}, u_{Hopf}$ and $u_{final}$ are neuron variables, from which at any given time only one of those neurons is non-zero.
The Hopf oscillator is formulated as follows:

\begin{align}
\label{eq:21}
f_{Hopf}(a-R_h, b) &= \left(
\begin{matrix}
\lambda & -\omega \\
\omega & \lambda
\end{matrix} \right)
\left(\begin{matrix} 
\alpha - R_h\\
b \nonumber
\end{matrix}\right)\\ &- \gamma [(a-R_h, b)^2 + b^2 ]
\left(\begin{matrix}
a-R_h\\b
\end{matrix}\right)
\end{align}

Depending on the angular frequency, the cycle is defined by $T = 2\pi/\omega$. The parameters $\lambda > 0 $ and $ \gamma > 0$ are controlling the limit cycle's radius.

\begin{equation}
\label{eq:23}
    R_h=\sqrt{\frac{\lambda}{\gamma}}
\end{equation}

Considering phase space, this limit cycle is shifted along the $\alpha$ axis by the radius $R_h$, so that the variable $\alpha$ (velocity) sinusoidally rises from $0$ to $2R_h$ and back to $0$ in one cycle.
After the completion of one cycle, the traveled length can be calculated with the following formula:

\begin{equation}
s(t)=\int_0^t \mathrm{R_h (1-cos(\omega\tau))d\tau=R_h\left(t- \frac{1}{\omega}sin(\omega t)\right)}
\end{equation}

\section{Implementation}
The methods described above were first tested by simulation and later implemented on an anthropomorphic robot arm (KUKA LBR iiwa 7 R800). Our intention was to use a robot arm for stopping a pendulum, similarly to how humans would interact. The whole project was first simulated in Unity (\autoref{fig:7}) including the moving pendulum and the robot system (apart from the ``Fast Robot Interface'', see below). 

\subsection{Experimental setup}
The robot KUKA LBR iiwa 7 R800 was used as robotic system. A tennis ball attached to a string of 1.93 meters length served as a pendulum. It was held by an electromagnet (2.5 kg payload, 3 watts) that can be controlled by software. To establish a connection between the magnet and the software, we used an Ardunio Uno R3 and its USB connection. Serial commands from the main program are sent and evaluated on the Arduino. Thus, depending on the command sent, the magnet is switched either on or off. In order to control the electromagnet with an Arduino, an additional circuit (based on a transistor as switch and a diode for overvoltage protection) is required, because each output of the Arduino may only be loaded with 5 mA.

\begin{figure}[htp]
    \centering
    \includegraphics[width=9cm]{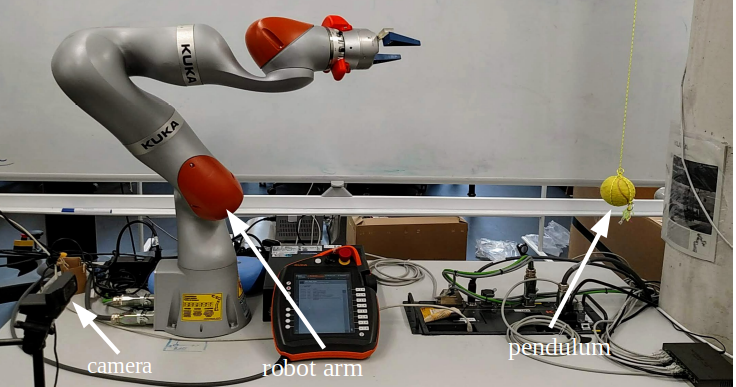}
    \caption{Experimental setup}
    \label{fig:5}
\end{figure}

\begin{figure}[htp]
    \centering
    \includegraphics[width=9cm]{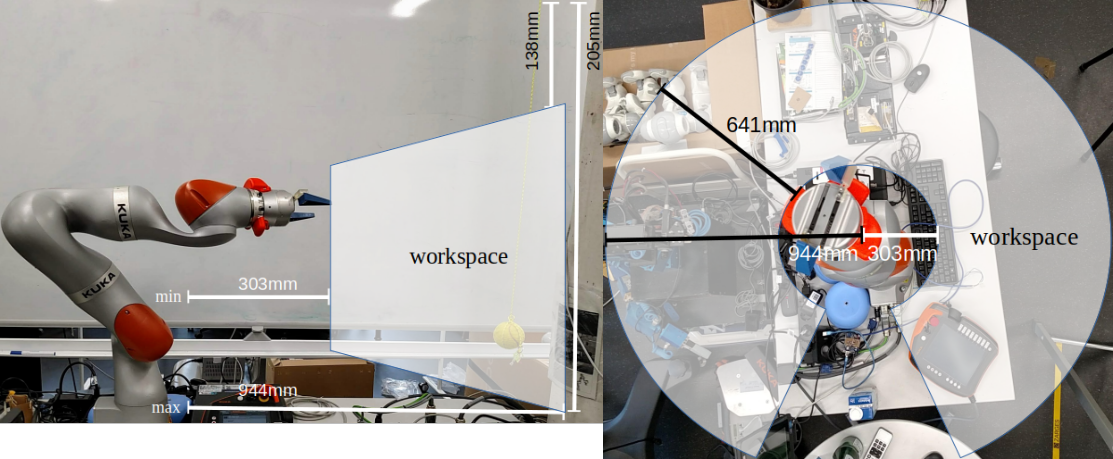}
    \caption{Workspace of robot}
    \label{fig:6}
\end{figure}

The camera (Logitech C920 HD Pro) is located next to the robot arm and is aligned parallel to the robot's direction of movement. The pendulum swings perpendicularly to the camera orientation.
The boundary conditions for the experiment are mainly set by the technical properties of the robot arm. The link lengths, which include the height of the base, the lengths of the forearm and upper arm and the length of the gripper determine the maximum reaching range of the arm. The smallest possible distance to the robot is given by the joint angle limits. These conditions result in the following working range of the robot arm (see \autoref{fig:5} and \autoref{fig:6}). 

\subsection{Simulation}
\begin{figure}[htp]
    \centering
    \includegraphics[width=8cm]{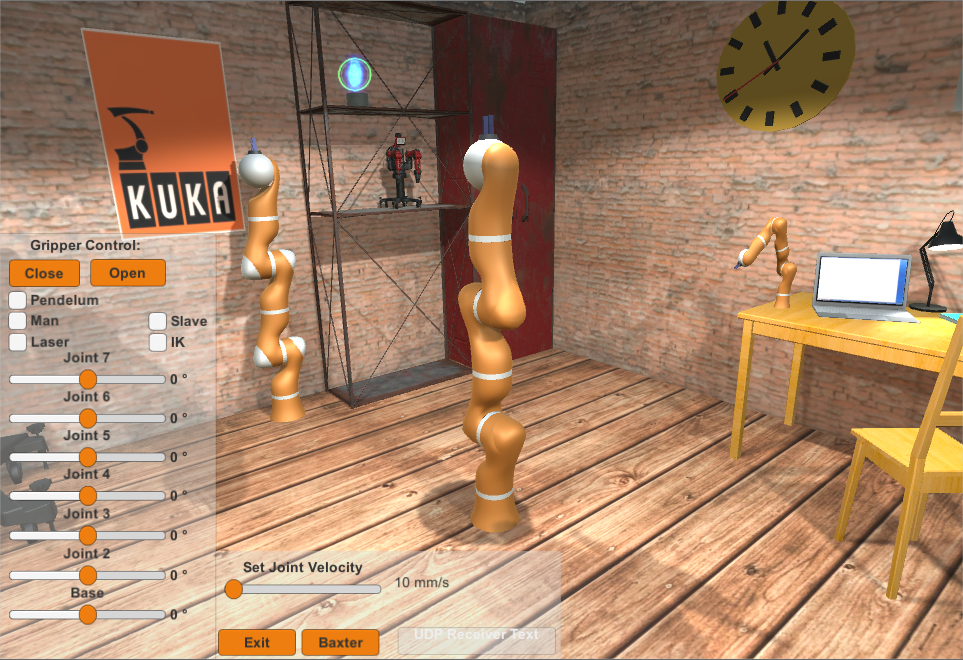}
    \caption{Simulation in Unity}
    \label{fig:7}
\end{figure}

In order to safely test and optimize the Kalman filter and the robot motion based on dynamic systems, a Unity simulation environment (\autoref{fig:7}) was created. The pendulum implemented here has no physics. The calculated deflection angle is sent to Unity via UDP and applied to the pendulum. Controlling the robot in Unity is done in the same way. Unity is used here only for simulation and has no control over the robot arm.

\subsection{Implementation of the movement on the real robot arm}
Kuka offers two interfaces for motion programming of the robot arm: the KUKA Line Interface (KLI) for non-real-time motion programming (not used here) and the KUKA Option Network Interface (KONI), also known as Fast Robot Interface or Fast Research Interface (FRI), for real-time motion programming. Real-time capable here means that clock timings from one to 15 milliseconds are possible.

\subsubsection{KUKA Line Interface (KLI)}
The KUKA Line Interface is designed for clock cycles of two milliseconds or more. The user has to specify the type of movement, the target position and the speed. The Sunrise Cabinet robot controller then automatically calculates and executes all the necessary parameter settings. The starting point of a movement is always the target point of the previous movement.
Since this interface has a ``starting'' latency of about 1 second, this type of connection is not chosen in this project

\subsubsection{KUKA Option Network Interface (KONI) / Fast Robot Interface (FRI)}
The task of the KUKA Option Network Interface is to monitor the robot status and to influence or superimpose robot movements. This interface allows for a continuous and real-time data exchange with the robot arm. Data, including movement commands and sensor information, can be exchanged in cycles of milliseconds.
Kuka has divided this interface into two parts. The FRI client application runs on the user's computer, while the robot application runs on the Sunrise Cabinet controller. The FRI channel is located between these two parts. The actual program for the path overlay is created in the FRI client application and if necessary changed in real-time.
As soon as a program with FRI functionality is started by the user, the robot controller is in monitor mode. There are two states: ``monitoring-wait'' and ``monitoring-ready''. In the first state, the robot controller has opened the FRI connection and is waiting for real-time data exchange with the FRI client application on the user's computer. In the second state, the robot controller performs real-time data exchange with the FRI client application. The robot controller can only switch to command mode (``commanding-active'') in the ``monitoring-ready'' state. The change is initiated in the robot application when a movement with path overlay is called up in the FRI client application.
The robot application is responsible for the robot's sequence control as well as the administration of coordinate systems, objects and movements. Additionally, the FRI channel is configured and managed, while the access from the FRI client application to the robot arm is enabled. The user is responsible to provide a trajectory, a velocity profile and the inverse kinematics. It should also be noted that a cycle in the FRI client application does not violate the real-time conditions (FRI clock), e.~g. by processing a time-consuming while loop. The implemented algorithm can be summarized by the following flowchart:

\begin{figure}[htp]
    \centering
    \includegraphics[width=8cm]{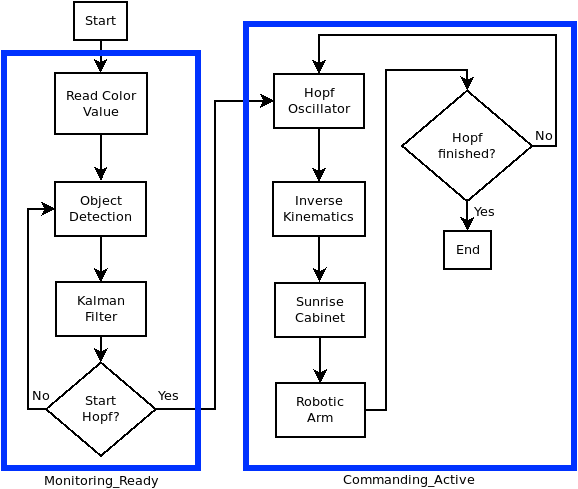}
    \caption{Flowchart of the project with real-time functionality. The time- and computationally intensive parts of the program were added to the step ``Monitoring\_Ready''. As soon as the interception condition or impact condition is fulfilled, the robot controller switches to ``Command\_Active''}
    \label{fig:8}
\end{figure}

\section{Experimental Results}
In the beginning of the flow chart as shown in \autoref{fig:8} stands the object detection for which a combination of color separation and shape recognition was used. Since the experiment takes place under static lab conditions, this algorithm resulted in a very reliable detection.

In the next step, the detected position was transferred to the EKF which increased the accuracy of measurement of the object's position furthermore and made it possible to predict its future position. Due to the intense tuning of the filter parameters conducted in advance in MATLAB, the object detection algorithm was well suited for this task. Nevertheless, there was still a small chance that in some recorded frames the ball's position could not be ascertained. To avoid this, the output of the EKF was used for the Hopf oscillator in the next processing step.

The Hopf oscillator was utilized in order to achieve a temporal stabilized motion. It started when the interception condition was fulfilled. Since its input arguments were only the distance and the desired time to reach the goal, this algorithm was very adaptive and could be adjusted to any trajectory.

After the first test runs, it turned out that the performance limits of the robot arm were reached very quickly. First the trajectory was running at lower speed. The speed was gradually increased until the executed trajectory deviated from the planned one. After several attempts at high speed, the overcurrent protection of the robot controller was activated. In order to avoid this, the speed was kept low and a linear trajectory was chosen. This required several attempts, as the occurring speeds and accelerations for each joint were difficult to estimate. 

The result was a slow, purposeful movement that was sufficient to intercept and stop the ball at the moving pendulum. It took about 1.5 seconds to reach the target from the initial position (distance of 0.70 meters). Unfortunately, this time was not constant and can change slightly with every reboot of the system. It was found out, that there are timing issues with the robot system itself, which could not be resolved because of very limited  hardware access. The 1.5 seconds was a good average assumption based on multiple experiments. 
With this assumption, 93\% of all tries a were precisely fulfilled.

However, an active stroke movement as an extension of the task could not be executed on this system due to the robot's lack of power reserves.

\section{Conclusion and Future Work}
This work successfully achieved to model the behaviour of temporally coordinated humanoid arm movement, precisely sequentially timed action sequences. The goal was to transfer this ability to a machine with specific focus on isochrony and temporal stabilization based on a Hopf oscillator. After detecting the object by a vision algorithm, measurement errors of the target's position were corrected and predicted  by an EKF. Afterwards, the inverse kinematics was calculated for the trajectory and the robot followed it based on the isochronously stabilized velocity profile generated by the Hopf oscillator. Due to latency issues inherent to the design of the robot, the speed of the target interception was kept moderate (0.70 meters in 1.5 seconds); thus, the system was reasonably responsive.

We successfully tested this setup for different workspace configurations, precisely distances from 0.35 to 0.80 meters. The maximum possible distance is affected by the robot's limb dimensions, while the respective minimum is constrained by the allowed joint angle values. The detection algorithm itself constitutes no bottleneck and works for a wider range of distances. The robot arm's lack of power to follow the trajectory is a shortcoming of the utilized system.

Nevertheless, future work could comprise to strike a ball with this robot arm. The ball would have to be thrown from a very long distance so that the arm has enough time to perform the appropriate trajectory execution. This could, for example, be a useful extension of the task. However, this would require a revision of the target recognition algorithm and the EKF system model. Furthermore, the presented approach should be evaluated on more powerful types of robotic systems.

Besides temporarily stabilized execution of repeatable and smooth movements with a sinusoidal velocity profile stemming from the Hopf oscillator, further aspects of biologically inspired movements should be incorporated to progressively converge towards humanoid motion. In this sense, this work could be combined with approaches considering the two-thirds power law or movement path planarity, for example.


%\section*{Appendix}

%Do we need one??

%%%%%%%%%%%%%%%%%%%%%%%%%%%%%%%%%%%%%%%%%%%%%%%%%%%%%%%%%%%%%%%%%%%%%%%%%%%%%%%%

\nocite{*}
%\bibliography{references,relatedWork}
\bibliography{relatedWork}
\bibliographystyle{plain}

\end{document}